%% file: main_final.tex
\documentclass{article}

    \newcommand{\arcsec}{^{\prime\prime}}
    \usepackage{xcolor}
    \usepackage[pdftex]{graphicx}

     \usepackage[nonatbib, final]{neurips_2019}

\usepackage[utf8]{inputenc} 
\usepackage[T1]{fontenc}    
\usepackage{deluxetable}
\usepackage{hyperref}       
\usepackage{url}            
\usepackage{booktabs}       
\usepackage{amsfonts}       
\usepackage{nicefrac}       
\usepackage{microtype}      
\usepackage{subcaption}

\title{Probabilistic Super-Resolution of Solar Magnetograms: 
Generating Many Explanations and Measuring Uncertainties}

\author{%
  Xavier Gitiaux\\
  George Mason University \\
  \texttt{xgitiaux@gmu.edu} \\
  \And
  Shane A.~Maloney \\
  Trinity College Dublin / \\
  Dublin Institute for Advanced Studies \\
  \texttt{shane.maloney@tcd.ie} \\
   \And
  Anna Jungbluth \\
  University of Oxford \\
  \texttt{anna.jungbluth@physics.ox.ac.uk}\\
  \And
  Carl Shneider \\
  Center for Mathematics and Computer Science (CWI) \\
  \texttt{carl.shneider@cwi.nl} \\
     \And
  Paul J.~Wright \\
  Stanford University \\
  \texttt{pjwright@stanford.edu} \\
   \And
  Atılım Güneş Baydin \\
  University of Oxford \\
  \texttt{gunes@robots.ox.ac.uk} \\
   \And
  Michel Deudon \\
  Element AI \\
  \texttt{michel.deudon@elementai.com} \\
  \And
   Yarin Gal \\
  University of Oxford \\
   \texttt{yarin@cs.ox.ac.uk} \\
    \And
 \And 
  Alfredo Kalaitzis \\
  Element AI \\
  \texttt{freddie@element.ai} \\
  \And
  Andr\'es Mu\~noz-Jaramillo \\
  Southwest Research Institute \\
  \texttt{amunozj@boulder.swri.edu} \\ 
}

\begin{document}

\maketitle

\begin{abstract}
Machine learning techniques have been successfully applied to super-resolution tasks on natural images where visually pleasing results are sufficient. However in many scientific domains this is not adequate and estimations of errors and uncertainties are crucial. To address this issue we propose a Bayesian framework that decomposes uncertainties into epistemic and aleatoric uncertainties. We test the validity of our approach by super-resolving images of the Sun's magnetic field and by generating maps measuring the range of possible high resolution explanations compatible with a given low resolution magnetogram.
\end{abstract}

\section{Introduction}

Deep learning has been successful at super-resolution (SR) of natural images, i.e. reconstructing  higher resolution (HR) images from low resolution (LR) inputs \cite{2018arXiv180803344Y}. However, to the extent of our knowledge, there is no existing work measuring uncertainty in super-resolution tasks. For scientific applications, estimating the uncertainty of a SR output is as important as the prediction itself, all the more as super-resolution is an ill-posed problem with many super-resolved images being consistent with the same low-resolution input.

To obtain robust uncertainties, we propose a Bayesian framework as in Kendall \& Gal \cite{kendall2017uncertainties} that decomposes uncertainty into {\it epistemic} and {\it aleatoric} uncertainty. {\it Epistemic} uncertainty relates to our ignorance of the true data generating process, and {\it aleatoric} uncertainty captures the inherent noise in the data. We apply our framework to images of the Sun's derived magnetic field (magnetograms), which are used to study the solar corona \cite{linker1999magnetohydrodynamic} and to predict space-weather events \cite{toth2005space}. These applications often require magnetograms spanning time-ranges longer than the lifetime of any single instrument. SR can compensate for inhomogeneities and discontinuities between instruments by converting multiple surveys to a single common resolution (e.g. \cite{2018A&A...614A...5D}).

In practice, we convert a state of the art super-resolution encoder-decoder architecture, HighRes-net \cite{anonymous2020highresnet}\footnote{\url{https://github.com/ElementAI/HighRes-net}.}, into a Bayesian deep learning framework by adding dropout at each convolutional layer, and tracking both the mean and variance of magnetic field values. We test the effectiveness of our framework by super-resolving magnetograms from the Helioseismic and Magnetic Imager \cite{2012SoPh..275..207S}. 

\begin{figure}[!h]
    \centering
    \begin{subfigure}{.49\textwidth}
    \includegraphics[width=1\textwidth]{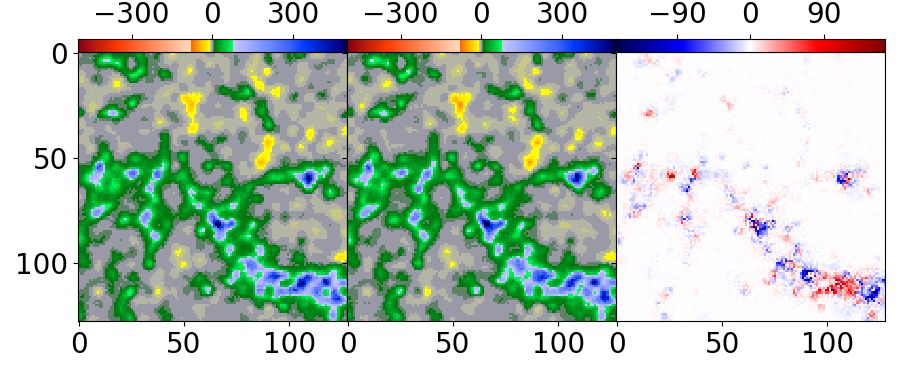}
    \caption{}
    
    \end{subfigure}
     \begin{subfigure}{.49\textwidth}
    \includegraphics[width=1\textwidth]{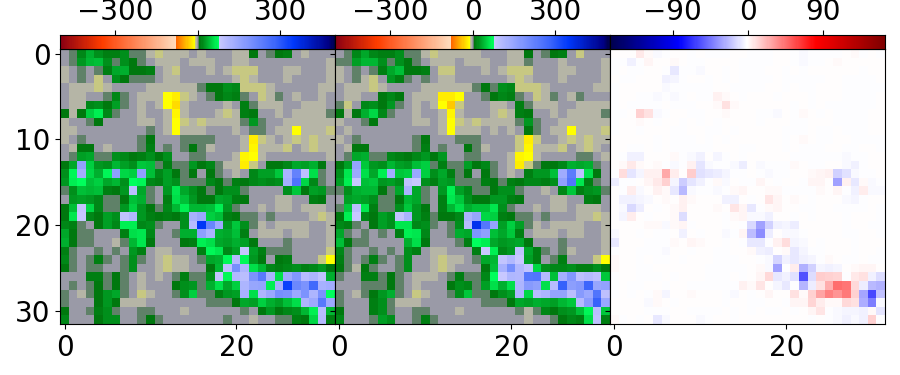}
      \caption{}
    \end{subfigure}
\caption{
Comparison of two realisations from HighRes-net, trained with dropout and heteroskedastic loss (see section 2), obtained from the same input using dropout during inference. HR models outputs and differences ({\it a}) and HR outputs degraded to match LR input ({\it b}). The differences are clearly visible in the HR ({\it a}) but reduced in the LR, which illustrates how many model's explanations might be consistent with the same low-resolution input.
}
\label{ref:fig1}
\end{figure}

We demonstrate that modelling {\it epistemic} uncertainty allows us to measure the range of high resolution explanations consistent with a low-resolution input. Figure \ref{ref:fig1} illustrates how our super-resolution architecture can generate two different extrapolations of the same input and how the difference is more attenuated once downsampled\footnote{Downsampling consists in two operations: (i) smoothing that convolves the high resolution magnetogram with a Gaussian kernel; (ii) downsampling that averages magnetic fields on each block of the size of the downscaling factor.}. Moreover, high resolution explanations have larger variance in regions of the Sun with a large magnetic field (so called active regions). We show that this larger variance cannot be properly accounted for unless disentangled from the larger amount of noise in regions with larger magnetic fields. In fact, {\it aleatoric} uncertainty is a full order of magnitude larger in active regions.

\section{Quantifying Uncertainties in Super-resolution}

If an image $y$ is downsampled by an unknown transformation into a LR image $x$, SR consists of learning the inverse transformation $f$ from $x$ to $y$. This is an ill-posed problem as many SR outputs map to the same LR input. Despite this difficulty, deep learning architectures can achieve SR, obtaining  $y$ by parameterizing $f$ with an encoder-decoder neural network $f_{w}(x)$. 
To model {\it epistemic} uncertainty we use a Bayesian framework from Kendall \& Gal \cite{kendall2017uncertainties} and impose a prior distribution on the model parameters $w$. Given a sample $\mathcal{D}_{n}=\{(x_{1}, y_{1}), ..., (x_{n}, y_{n})\}$ of magnetograms, we evaluate the posterior distribution $p(w|\mathcal{D}_{n})$ by Bayesian inference. In practice, this inference is approximated by dropout variational inference \cite{gal2015bayesian}, i.e. adding dropout to each layer during training and inference to sample a series of SR realizations.

{\it Aleatoric} uncertainty is modelled by assigning a distribution to the outputs of the model $p(y|f_{w}(x))$. We assume that for each pixel $j$, the noise is Gaussian with a variance $\sigma_{j}(x)^{2}$ that depends on the input image and pixel position $j$ in the detector. We chose to model this heteroskedascity, i.e. the variation of {\it aleatoric} uncertainty across pixels, because projection distortions are a function of pixel position and conditional on a low-resolution magnetic field, the distribution of magnetic field is noisier in active regions of the Sun (Figure \ref{fig:lr-hr}). 

Assuming that pixel noise is independent across images and pixels, the negative log-likelihood of a sample $\mathcal{D}_{n}$ is
\begin{equation}
\label{eq: 1}
    \displaystyle\sum_{i=1}^{n}\sum_{j=1}^{m}\frac{1}{2\sigma_{j}(x_{i})^{2}}(y_{ij} - f_{w, j}(x))^{2} + \frac{1}{2}\log(\sigma_{j}(x_{i})^{2}) \, ,
\end{equation} 

where $m$ is the number of pixels in each high resolution image. To estimate both {\it epistemic} and {\it aleatoric} uncertainties, we use a Bayesian neural network that outputs both $f_{w}(.)$ and $\sigma(.)$ and is trained to minimize the heteroskedastic loss (Equation \ref{eq: 1}). For an input image $x$, the magnetic field uncertainty is obtained by sampling $T$ times the network weights, getting $f_{t}(x)$ and $\sigma_{t}(x)^{2}$ for each sample $t$ and computing predictive uncertainty as

\begin{equation}
\underbrace{\frac{1}{T}\displaystyle\sum_{t=1}^{T}f_{t}(x) ^{2} - \left(\frac{1}{T}\displaystyle\sum_{t=1}^{T}f_{t}(x)\right)^{2}}_{Epistemic} + \underbrace{\frac{1}{T}\displaystyle\sum_{t=1}^{T}\sigma_{t}(x) ^{2}}_{Aleatoric} \, .
\end{equation}

\section{Experiments}
\label{data_arch}
\subsection{Data and Architecture}
We use $400,000$ ($4096 \times 4096$ pixel) magnetograms collected by HMI between 2010 and 2019. 
Our experiment consists of super-resolving HMI magnetograms that have been artificially reduced by a factor of 4 in resolution by smoothing the HR image with a Gaussian kernel before down-sampling by averaging the magnetic fields of each 2 by 2 block of the HR magnetogram.
The data is split into training, validation, and test sets by allocating one month randomly drawn every year to each of the test and validation sets and the remaining months to the training set. 

We conduct this experiment with a state-of-the-art super-resolution architecture, HighRes-net \cite{anonymous2020highresnet}\footnote{\url{https://github.com/ElementAI/HighRes-net}.} which we adapted to output both mean and variance of each pixel value in high resolution. Dropout is modelled at each layer by a Bernouilli distribution with parameter $p=0.2$. We run a set of experiments on $128 \times 128$~pixel patches cropped from the center of HR images. Variational inference is done by sampling $500$ realizations of the model. 

\subsection{Results}

We test how including epistemic and aleatoric uncertainty affects the accuracy of the SR task. Table~\ref{tab:1} compares the mean squared error (MSE) of our baseline, HighRes-net trained with a MSE loss to (i) HighRes-net trained with dropout and MSE loss; (ii) HighRes-net trained with a heteroskedastic loss (Equation~\ref{eq: 1}); and, (iii) HighRes-net trained with both dropout and heteroskedastic loss. Although accounting separately for {\it epistemic} and {\it aleatoric} uncertainty does not seem to degrade the performance of the neural network, the MSE is $8\%$ higher with a combination of aleatoric and epistemic uncertainty. We found that the model with combined uncertainty is sensitive to the initialization of heteroskedastic variances. In our current implementation, we compute pixel j's heteroskedastic variance as $\sigma_{j}^{2}(x) = \tilde{\sigma_{j}}^{2} + \overline{\sigma}^{2}$, where $\overline{\sigma}^{2}$ is the mean squared error obtained with homoskedastic variances. However, more work is needed to improve the model performance

\input{NeuRIPS2019/tables/table1.tex}

Modelling  {\it epistemic} uncertainty allows us to capture that the super-resolution of magnetograms is a more ill-posed problem in active regions of the Sun with large magnetic fields. Figure \ref{ref:fig1} shows that predictions consistent with the same low-resolution input are more likely to disagree for large magnetic fields. A possible explanation is the lack of examples of active regions, particularly around the solar Equator. 

\begin{figure}
    \centering
    \includegraphics[width=0.95\textwidth]{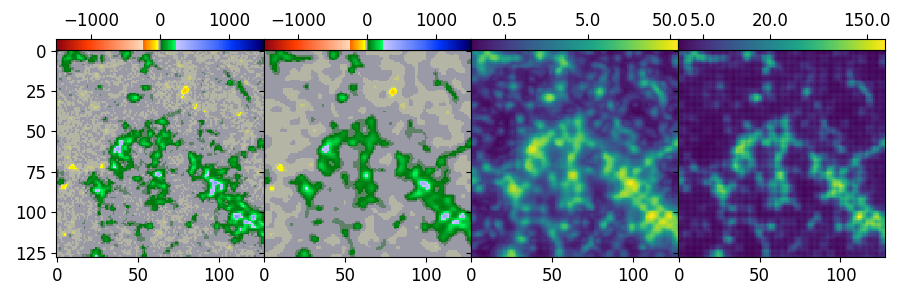}
    \caption{({\it left to right}) An example of a HR target image ($128\arcsec \times 128\arcsec$) plotted over $\pm1500$ Gauss; the corresponding mean of MC-dropout samples; model uncertainty;  and estimated noise.}
    \label{fig:mc_drop_images}
\end{figure}

However, this lack of data is confounded by the presence of more noise in active regions: Figure \ref{fig:lr-hr} shows that the distribution of magnetic fields in high resolution has higher variance conditional on its counterpart value in low-resolution. {\it Aleatoric} uncertainty effectively measures the large amount of noise in regions with a large magnetic field (Figure \ref{fig:mc_drop_images} and \ref{fig:mc_drop_images_all}). By accounting for both {\it aleatoric} and {\it epistemic} uncertainty,  we can properly disentangle the ill-posed nature of the super-resolution task from the heteroskedasticity of the noise (Figure \ref{fig:mc_drop_images} and \ref{fig:mc_drop_images_all}). We observe that in active regions, {\it aleatoric} uncertainty is larger than {\it epistemic} uncertainty. 

\section{Conclusions and Future Work}
We employ a Bayesian deep learning framework as a way to measure how uncertain predictions of super-resolved magnetograms are, particularly in active regions of the Sun. The task is uniquely challenged by the confounding effect of larger noise in active regions and relative scarcity of regions with large magnetic fields. We model {\it aleatoric} and {\it epistemic} uncertainty to properly measure the range of high resolution explanations compatible with a given low resolution input. This work is a first step toward generating super-resolved magnetograms useful to the heliophysics community. Future avenues for research include (i) expanding our results to the full solar disk; and, (ii) generating an initialisation procedure that improves the accuracy of our Bayesian model.

\subsubsection*{Acknowledgments}
This work was conducted at the NASA Frontier Development Laboratory (FDL) 2019. NASA FDL is a public-private partnership between NASA, the SETI Institute and private sector partners including Google Cloud, Intel, IBM, Lockheed Martin, NVIDIA, and Element AI. These partners provide data, expertise, training, and compute resources necessary for rapid experimentation and iteration in data-intensive areas.
P.~J.~Wright acknowledges support from NASA Contract NAS5-02139 (HMI) to Stanford University. This research has made use of the following open-source Python packages SunPy \cite{sunpy}, NumPy \cite{numpy}, Pandas \cite{pandas}, and PyTorch \cite{pytorch}. We thank Santiago Miret and Sairam Sundaresan (Intel) for their advice on this project. 

\bibliographystyle{unsrt}
\bibliography{main_final.bbl}

\newpage
\appendix
\section{Appendix}

\begin{figure}[!h]
    \centering

    \includegraphics[width=0.95\textwidth]{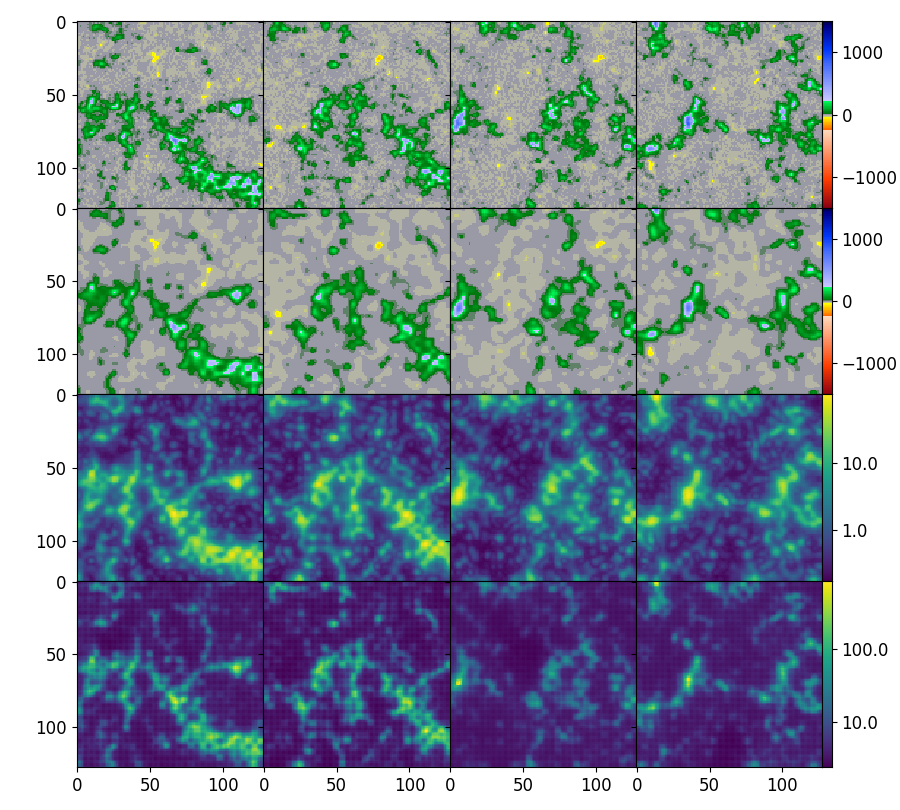}
    \caption{Sample of four HR target images ($128\arcsec \times 128\arcsec$) plotted over $\pm1500$ Gauss ({\it 1\textsuperscript{st} row}) the corresponding mean of MC-dropout samples ({\it 2\textsuperscript{nd} row}) model uncertainty ({\it 3\textsuperscript{rd} row})  and estimated noise ({\it 4\textsuperscript{th} row}).}
    \label{fig:mc_drop_images_all}
\end{figure}

\begin{figure}[!h]
    \centering
    \includegraphics[width=0.95\textwidth]{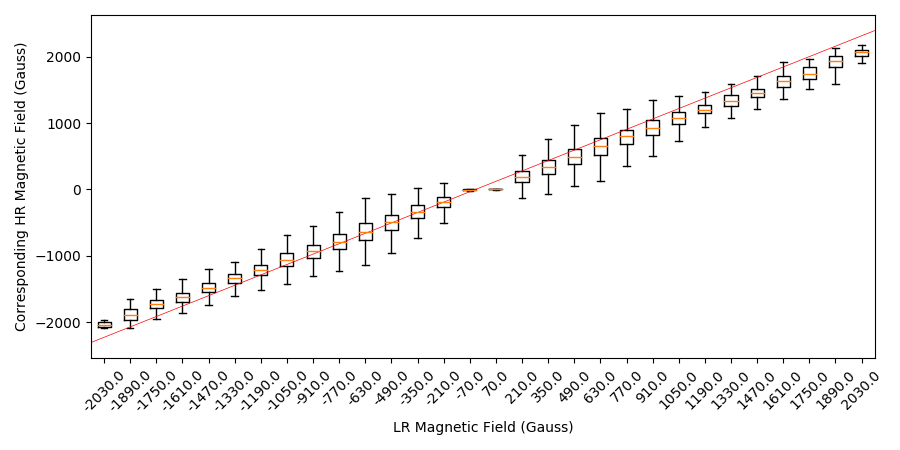}
    \caption{({\it left}) Empirical mapping between single full disk LR and HR magnetorams. There is a clear increase in the difference of the mean and variance as a function of the LR field strength.
    }
    \label{fig:lr-hr}
\end{figure}{}
\medskip

\small

\end{document}

%% file: NeuRIPS2019/tables/table1.tex
\begin{table}[t!]
\centering
 \begin{tabular}{ll} 
 \hline
 Models & MSE \\ 
 \hline
 \hline
 HighRes-net & 88.45  \\
 + Epistemic & 90.00  \\
 + Aleatoric & 90.47  \\
 + Epistemic \& Aleatoric & 98.10  \\
 [1ex] 
 \hline
 \end{tabular}
 \vspace{2ex}
 \caption{Super-resolution performance: Mean Squared Error of HighRes-net with and without dropout and heteroskedastic loss.  Lower is better. This shows that accounting for aleatoric or epistemic uncertainty separately while training HighRes-net does not degrade the model accuracy. However, the MSE is $8\%$ higher with a combination of aleatoric and epistemic uncertainty} \label{tab:1}
\end{table}